\documentclass[runningheads]{llncs}
\usepackage{graphicx}
\usepackage{amsmath}
\usepackage{cases}
\usepackage{multirow}
\usepackage{amsmath}
\usepackage{booktabs}
\usepackage{floatrow}
\usepackage{marvosym}

\newfloatcommand{capbtabbox}{table}[][\FBwidth] 
\begin{document}
\title{DSKG: A Deep Sequential Model for\\ Knowledge Graph Completion}

\author{Lingbing Guo\inst{1,2} \and Qingheng Zhang\inst{1} \and Weiyi Ge\inst{2} \and
Wei Hu\inst{1}\textsuperscript{(\Letter)} \and Yuzhong Qu\inst{1} }
\authorrunning{L. Guo et al.}

\institute{
	State Key Laboratory for Novel Software Technology, Nanjing University, China \\
	\email{\{lbguo, qhzhang\}.nju@gmail.com, \{whu, yzqu\}@nju.edu.cn} \and
	Science and Technology on Information Systems Engineering Lab, Nanjing, China\\
	\email{geweiyi@163.com}
}
\maketitle              % typeset the header of the contribution
\pagestyle{empty}
\begin{abstract}
Knowledge graph (KG) completion aims to fill the missing facts in a KG, where a fact is represented as a triple in the form of $(subject, relation, object)$. Current KG completion models compel two-thirds of a triple provided (e.g., $subject$ and $relation$) to predict the remaining one. In this paper, we propose a new model, which uses a KG-specific multi-layer recurrent neural network (RNN) to model triples in a KG as sequences. It outperformed several state-of-the-art KG completion models on the conventional entity prediction task for many evaluation metrics, based on two benchmark datasets and a more difficult dataset. Furthermore, our model is enabled by the sequential characteristic and thus capable of predicting the whole triples only given one entity. Our experiments demonstrated that our model achieved promising performance on this new triple prediction task.

\keywords{knowledge graph completion \and deep sequential model \and recurrent neural network}
\end{abstract}

\section{Introduction}

\begin{figure}[t]
	\centering
	\includegraphics[width=.75\columnwidth]{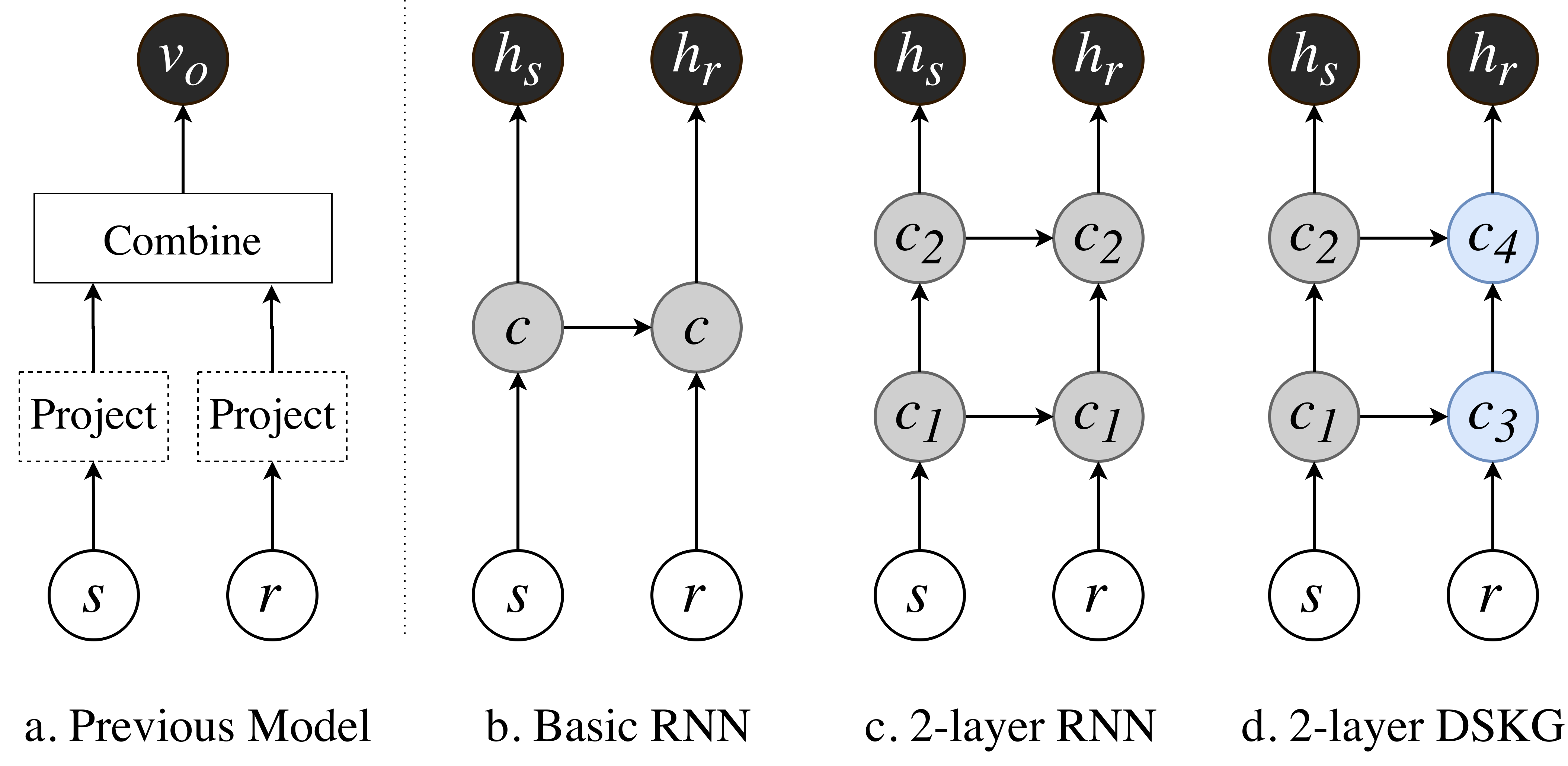}
	\caption{Different models for entity prediction. White and black circles denote input and output vectors, respectively. $c$ denotes an RNN cell and $h$ denotes a hidden state. DSKG uses $c_1, c_2$ to process entity $s$, and $c_3, c_4$ to process relation $r$. All of them are different RNN cells.}
	\label{fig:methods}
\end{figure}

Knowledge graphs (KGs), such as Freebase \cite{Freebase} and WordNet \cite{WordNet}, typically use triples, in the form of $(subject, relation, object)$ (abbr. $(s, r, o)$), to record billions of real-world facts, where $s,o$ denote entities and $r$ denotes a relation between $s$ and $o$. Since current KGs are still far from complete, the KG completion task makes sense. Previous models focus on a general task called entity prediction (a.k.a. link prediction) \cite{TransE}, which asks for completing a triple in a KG by predicting $o$ (or $s$) given $(s, r, ?)$ (or $(?, r, o)$). Fig.~\ref{fig:methods}a shows an abstract model for entity prediction. Input $s,  r$ are firstly projected by some vectors or matrices, and then combined to a continuous representation $v_o$ to predict $o$.

Although previous models perform well on entity prediction, they may still be inadequate to complete a KG. Let us assume that a model can effectively complete an entity $s$ given a relation $r$ explicitly. If we do not provide any relations, this model is incompetent to fill $s$, because it is incapable of choosing which relation to complete this entity. Actually, the underlying data model of KGs does not allow the existence of any incomplete tuple $(s,r)$. 

The recurrent neural network (RNN) is a neural sequence model, which has achieved good performance on many natural language processing (NLP) tasks, such as language modeling and machine translation \cite{LanguageModel,Translation}. Triples in a KG can be approximately regarded as simple sentences of length 3. For example, a triple $(\textit{USA, contains, NewYorkCity})$ can be transformed to a sentence \textit{``USA contains New York City"}. This enlightens us to use RNNs to model KGs. However, we are still challenged by the following problems: (i) Triples are not natural language. They model the complex structure with a fixed expression $(s, r, o)$. Such short sequences may be insufficient to provide enough context for prediction. Meanwhile, it is costly and difficult to construct valuable long sequences from massive paths; (ii) Relations and entities are two different types of elements that appear in triples in a fixed order. It is inappropriate to treat them as the same type.

To solve the aforementioned problems, we propose DSKG (Deep Sequential model for KG), which employs RNNs with a new structure and uses a KG-specific sampling method for training. We design a basic RNN model as the initial version to illustrate our idea (see Fig.~\ref{fig:methods}b). This basic model takes input $s, r$ as the same type of elements and recurrently processes them. An RNN cell is denoted by $c$, which receives its previous hidden state and the current element as input to predict the next. The cell in the entity layer processes entities like $s$, while the cell in the relation layer processes relations like $r$. In this model, there only exists one cell to sequentially process all input elements, so $s, r$ are fed to the same cell $c$ to obtain their respective output. We then use $h_s$ to predict the relations of $s$ and $h_r$ to predict the objects of $s \to r$. 

The basic RNN model may be not model complex structures well, because it only uses a single RNN cell to process all input sequences. In the NLP area, researchers usually stack multiple RNN cells to improve the performance. Here, we borrow this idea to build a multi-layer RNN model (see Fig.~\ref{fig:methods}c). But still, this model cannot precisely model triples in a KG, since entities and relations have very different characteristics. 

As depicted in Fig.~\ref{fig:methods}d, the proposed DSKG employs respective multi-layer RNNs to process entities and relations. Specifically, DSKG uses independent RNN cells for the entity layer and the relation layer, i.e., $c_1, c_2, c_3, c_4$ in the figure are all different RNN cells. We believe that this KG-specific architecture can achieve better performance when relations are diverse and complex. Because DSKG considers predicting entities (or relations) as a classification task, we also propose a customized sampling method that samples negative labels according to the type of current training label. Furthermore, DSKG has the ability to predict the relations of one entity, which inspires us to employ a method to enhance the results of entity prediction by relation prediction. For example, when predicting $(\textit{USA, contains, ?})$, the model can automatically filter entities like people or movies, because these entities are not related to relation $\textit{contains}$. 

We conducted the entity prediction experiment on two benchmark datasets. The results showed that DSKG outperformed several state-of-the-art models in terms of many evaluation metrics. Furthermore, we evaluated DSKG on FB15K-237 \cite{Node+LinkFeat}. The results demonstrated that DSKG outperformed the other models. Additionally, we designed a new KG completion experiment for triple prediction, as a complement to entity prediction. We demonstrated that, as compared with the general multi-layer RNN models, DSKG also achieved superior results. Our source code, datasets and experimental results are available online\footnote{https://github.com/nju-websoft/DSKG}.

\section{Related Work}
\label{sect:work}

\subsection{TransE-like Models}

TransE \cite{TransE} represents entities and relations as $k$-dimensional vectors in a unified space, and models a triple $(s, r, o)$ as $s + r \approx o$. TransE works well for the one-to-one relationship, but fails to model more complex (e.g., one-to-many) relationships. TransH \cite{TransH} tries to solve this problem by regarding each relation $r$ as a vector on a hyperplane whose normalization vector is $w_r$. It projects entities $s,o$ to this hyperplane by $w_r$. TransR \cite{TransR} uses relation-specific matrices to project entities. For each relation $r$, it creates a matrix $W_r$ to project $s,o$ by $W_r$. TransR also adopts the same energy function. 

PTransE \cite{PTransE} leverages additional path information for training. For example, if there exist two triples $(e_1, r_1, e_2), (e_2, r_2, e_3)$, which can be regarded as a path in a KG, and another triple $(e_1, r_x, e_3)$ holds simultaneously, then $e_1 \to r_1 \to e_2 \to r_2 \to e_3$ is a valuable path and recorded as $(e_1, r_1, r_2, e_3)$. However, preparing desirable paths requires to iterate over all possible paths, thus this process may consume enormous resources for large KGs. Consequently, PTransE and other path-based models may be inefficient to model large KGs.

All the aforementioned models choose to minimize an energy function that is used in or similar to TransE. Moreover, TransR and PTransE use pre-trained entity and relation vectors from TransE as initial input.

\subsection{Other Models}

Some models are different from the TransE-like models. DISTMULT \cite{DISTMULT} is as simple as TransE, but uses a completely different energy function. It is based on a bilinear model \cite{Bilinear}, and represents each relation as a diagonal matrix. ComplEx \cite{ComplEx} extends DISTMULT with the complex embedding technique.

Node+LinkFeat (in short, NLFeat) \cite{Node+LinkFeat} can also be regarded as a path-based model similar to PTransE, but only needs to extract paths of length 1 for constructing node and link features. Although it uses paths of length 1, it still consumes considerable resources for large KGs. NeuralLP \cite{NeuralLP} aims at learning probabilistic first-order logical rules for KG reasoning. For each relation $r$, it creates an adjacency matrix $M^r$, where the value of $M^r_{i,j}$ is non-zero if triple $(e_i, r, e_j)$ exists. Then, NeuralLP learns to reason by conducting matrix multiplication among different adjacency matrices.

Recently, ConvE \cite{ConvE} combines input entity $s$ and relation $r$ by 2D convolutional layers. It first reshapes the vectors of input $s, r$ to 2D shapes, and then concatenates the two matrices for convolution operation. ConvE also describes a very simple model called InverseModel, which only learns inverse relations in a KG  but achieves pretty good performance. Other models like \cite{SSP,DKRL} use extra resources that cannot be obtained from the original training data, such as text corpora or entity descriptions. We do not consider them in this paper.

To the best of our knowledge, all the existing models require both one entity $s$ and its relation $r$ provided to complete a triple. The proposed model in this paper is the first work that can predict the whole triples only given $s$.

\section{Methodology}
\label{sect:method}

In this section, we first describe RNN and its multi-layer version. Then, we present DSKG, a variant of the multi-layer RNN specifically designed for KGs. To train DSKG effectively, we also propose a type-based sampling method. Finally, we introduce a method to enhance entity prediction with relation prediction.

\subsection{RNN and Its Multi-layer Version}

We start with the basic RNN model, which has only one RNN cell. Given a sequence $(x_1, \ldots, x_T)$ as input, the basic RNN model processes it as follows:
\begin{equation}
h_t = f(W_h h_{t-1} + W_x x_t + b),
\label{eq:rnn}%
\end{equation}
where $f(\cdot)$ is an activation function, and $W_h, W_x, b$ are parameters. $h_t$ is the output hidden state at timestep $t$.

Multi-layer RNNs have shown promising performance on modeling complex hierarchical architectures in the NLP area \cite{DRNN}. By stacking multiple RNN cells, the complex features of each element can be hierarchically processed (see Fig.~\ref{fig:methods}c). We write this below: 
\begin{equation}
	h_t^i = \begin{cases}
		f(W_h^i h_{t-1}^i + W_x^i x_t + b^i) 	      & i=0 \\
		f(W_h^i h_{t-1}^i + W_x^i h_t^{i-1} + b^i) & i>0 \\
	\end{cases},
	\label{eq:drnn}%
\end{equation}
where $W_h^i, W_x^i, b^i$ are parameters for the $i$-th RNN cell. $h_t^i$ is the hidden state of the $i$-th RNN cell at timestep $t$. Hence, each input element would be sequentially processed by each cell, which can be regarded as combining the concept of deep neural network (DNN) with RNN. In the end, we can use the hidden state of the last cell as output $h_t$ at timestep $t$.

\subsection{The Proposed Deep Sequential Model}

Regarding triples in a KG as sequences enables us to model the KG with RNN. However, these length-3 sequences (i.e., triples) are quite special: entities and relations have very different characteristics and always crisscross in each triple. Therefore, we think that constructing respective multi-layer RNNs for entities and relations can help the model learn more complex structures. According to this intuition, we propose a KG-specific multi-layer RNN, which uses different RNN cells to process entities and relations respectively. As illustrated in Fig.~\ref{fig:methods}d, using this architecture, the entire network is actually non-recurrent but still sequential. We write this structure as follows:
\begin{align}
W_h^i = \begin{cases}
E_h^i \\
R_h^i 
\end{cases}
W_x^i = \begin{cases}
E_x^i  \\
R_x^i 
\end{cases}
b^i = \begin{cases}
b_E^i & x_t \in \mathbf{E}\\
b_R^i & x_t \in \mathbf{R}\\
\end{cases},
\end{align}
\label{eq:res}%
where $\mathbf{E}, \mathbf{R}$ denote the entity set and the relation set, respectively. We choose the current multi-layer RNN by the type of $x_t$, and then apply Eq. \eqref{eq:drnn} for calculation.

\subsection{Type-based Sampling}

Sampled softmax \cite{SampledSoftmax} is a very popular method for large label space classification. The underlying idea of this method is to sample a small number of negative classes to approximate the integral distribution. We write it as follows:
\begin{subequations}
	\begin{align}
	p_t &= W_o h_t + b_o,\\
	L_t &= -I(p_t, y_t) + \log\Big(\sum_{\tilde{y}\in \{y_t\}\cup \textbf{NEG}_t }^{} {\exp\big(I(p_t, \tilde{y})\big)} \Big),
	\end{align}
	\label{eq:sample}%
\end{subequations}
where $W_o, b_o$ are output weight matrix and bias. $I(p_t, y_t)$ returns the $y_t$-th value of $p_t$. We first use a fully-connected layer to convert output hidden state $h_t$ to an unscaled probability distribution of label space, and then carry out the sampled softmax method to calculate the cross-entropy loss $L_t$. \textbf{NEG}$_t$ denotes the negative set at timestep $t$. It is usually generated by a log-uniform sampler.

Furthermore, in KGs, label $y_t$ also has its type. When $y_t$ refers to an entity, it is meaningless to use negative relation labels for training, and vice versa. Therefore, we propose a customized sampling method that samples negative labels according to the type of $y_t$. We write it as follows:
\begin{equation}
\textbf{NEG}_t = \begin{cases}
Z(\mathbf{E}, n_e) & y_t \in \mathbf{E}\\
Z(\mathbf{R}, n_r) & y_t \in \mathbf{R}\\
\end{cases},
\label{eq:neg}%
\end{equation}
where $Z(\mathbf{E}, n_e)$ denotes the log-uniform sampler that samples the number of $n_e$ labels from entity set $\mathbf{E}$. $Z(\mathbf{R}, n_r)$ is defined analogously. It is worth noting that, this sampler needs the labels in a lexicon sorted in descending order of frequency, thus we should also separately calculate the frequencies of entities and relations.

\subsection{Enhancing Entity Prediction with Relation Prediction}
\label{subsect:enhance}

Due to the input is length-3 triples, the model only minimizes two sub-losses for each triple. Given a triple $(s, r, o)$, the model learns to predict $r$ based on $s$, and to predict $o$ based on $s \to r$. We propose a method that can leverage relation prediction for enhancing entity prediction. In Section~\ref{subsect:alternative}, the experimental analysis proves that learning to predict relations is helpful for entity prediction. 

Reversing relations is a commonly-used method to enable KG completion models to predict head and tail entities in an integrated fashion  \cite{PTransE,Node+LinkFeat}. Specifically, for each triple $(s, r, o)$ in the training set, a reverse triple $(o, r^-, s)$ is constructed and added into the training set. Thus, a model can predict tail entities with input $(s, r, ?)$, and predict head entities with $(o, r^-, ?)$.

Previous models for KG completion need $s, r$ to predict $o$. However, the ability of predicting relations enables our model to evaluate the probability distribution of reverse relations for each entity. For example, given an entity $e_j$, if the probability of $e_j \to r^-$ is very close to zero, then we can speculate that $e_j$ does probably not have the relation $r^-$. In other words, $e_j$ is not an appropriate prediction for $(s, r, ?)$. We formulate this by the following equation:
\begin{equation}
p_{(s, r,?)}^\prime = (p_{(\mathbf{E}, r^-)})^\alpha p_{(s, r, ?)},
\end{equation}
where $p_{(\mathbf{E}, r^-)}$ denotes the probability vector of $r^-$ for all entities, and we calculate its element-wise power of $\alpha$. We set $\alpha < 1$, since we want to alleviate the influence of such inaccurate prediction results. $p_{(s, r, ?)}$ denotes the original probability vector of $(s, r, ?)$. For example, assume that the original entity probability vector is $(0.25, 0.25, 0.25)$. If we set $\alpha =\frac{1}{3}$, a reverse relation probability vector $(0.001, 0.8, 0.9)$ would be refined to $(0.1, 0.93, 0.97)$. By element-wise multiplication of the original entity probability vector and the refined reverse relation probability vector, we have $(0.025, 0.233, 0.243)$, which slightly affects those entities with high probabilities of $r^-$, but seriously penalizes those entities with near-zero probabilities. Consequently, the differences between entity probabilities are enlarged to help predict entities more accurately.

\begin{table}[t]
	\centering	
	\caption{Entity prediction results on two benchmark datasets}
	\label{tab:ent}
	{\scriptsize \begin{tabular}{lcccccccc}
			\toprule
			\ \multirow{2}{*}{Models}\ & \multicolumn{4}{c}{FB15K} & \multicolumn{4}{c}{WN18} \\
			\cmidrule(lr){2-5} \cmidrule(lr){6-9} &\ Hits@1\ &\ Hits@10\ &\ MRR\ &\ MR\ &\ Hits@1\ &\ Hits@10\ &\ MRR\ &\ MR\ \\ \midrule
			\ TransE$^\dagger$\hfill \cite{TransE}\ & 30.5 & 73.7 & 45.8 & 71 & 27.4 & 94.4 & 57.8 & 431 \\
			\ TransR$^\dagger$\hfill \cite{TransR}\ & 37.7 & 76.7 & 51.9 & 84 & 54.8 & 94.7 & 72.6 & 415 \\
			\ PTransE$^\dagger$\hfill \cite{PTransE}\ & 63.8 & 87.2 & 73.1 & 59 & 87.3 & 94.2 & 90.5 & 516 \\ \midrule
			\ DISTMULT\hfill \cite{DISTMULT}\ & 54.6 & 82.4 & 65.4 & 97 & 72.8 & 93.6 & 82.2 & 902 \\
			\ NLFeat\hfill \cite{Node+LinkFeat}\ & - & 87.0 & \textbf{82.1} & - & - & 94.3 & 94.0 & - \\
			\ ComplEx\hfill \cite{ComplEx}\ & 59.9 & 84.0 & 69.2 & -  & 93.6 & 94.7 & 94.1 & - \\
			\ NeuralLP\hfill \cite{NeuralLP}\ & - & 83.7 & 76.0 & - & - & 94.5 & 94.0 & - \\
			\ ConvE\hfill \cite{ConvE}\ & 67.0 & 87.3 & 74.5 & 64 & 93.5 & 95.5 & 94.2 & 504 \\
			\ InverseModel\hfill \cite{ConvE}\ & 74.3 & 78.6 & 75.9 & 1,563 & 75.7 & \textbf{96.9} & 85.7 & 602 \\ \midrule
			\ DSKG (cascade)\ & 64.9 & 87.7 & 73.0 & 151 & 93.9 & 95.0 & 94.3 & 959 \\
			\ DSKG & \textbf{75.3}\ & \textbf{90.2} & 80.9 & \textbf{30} & \textbf{94.2} & 95.2 & \textbf{94.6} & \textbf{337} \\ \bottomrule
			\multicolumn{9}{l}{``$\dagger$" denotes the models executed by ourselves using the provided source code,} \\
			\multicolumn{9}{l}{\quad \; due to some metrics were not used in literature.} \\
			\multicolumn{9}{l}{``-" denotes the unknown results, due to they were unreported in literature and} \\
			\multicolumn{9}{l}{\quad \; we cannot obtain/run the source code.} \\
	\end{tabular}}
\end{table}

\section{Experiments}
\label{sect:exp}

\subsection{Datasets and Experiment Settings}

We implemented our model with TensorFlow and conducted a series of experiments on three datasets: FB15K \cite{TransE}, WN18 \cite{TransE} and FB15K-237 \cite{Node+LinkFeat}. Recent studies observed that FB15K and WN18 contain many inverse triple pairs, e.g., $(\textit{USA, contains, NewYorkCity})$ in the test set and $(\textit{NewYorkCity, containedby,}$ $\textit{USA})$  in the training set \cite{Node+LinkFeat}. By detecting the subjects and objects of $\textit{contains,}$ $\textit{containedby}$,  this inverse pair can be easily confirmed. So, the answer of $(\textit{USA,}$ $\textit{contains, ?})$ is exposed. Even a very simple model that concentrates on these inverse relations can achieve state-of-the-art performance for many metrics \cite{ConvE}.
Note that inverse triples are totally different from reverse triples. In our experiments, we more focus on FB15K-237, which was created by removing the inverse triples in FB15K. The detailed statistical data are listed in Table~\ref{tab:dataset}. We used the Adam optimizer \cite{Adam} and terminated training when the results on the validation data were optimized. For each dataset, we used the same parameters as follows: learning rate $\lambda=0.001$, embedding size $k=512$ (initialized with the xavier initializer), and batch size $n_B=2\text{,}048$. We chose the LSTM cells to implement the multi-layer RNNs and added the output dropout layer with keep probability $p_D=0.5$ for each cell. The main results reported in this section is based on the 2-layer DSKG model. We will show parameter analysis in Section~\ref{sect:analysis}. 

\subsection{Entity Prediction}

Following \cite{TransE,Node+LinkFeat,ConvE} and many others, four evaluation metrics were used: (i) the percentage of correct entities in ranked top-1 (Hits@1); (ii) in ranked top-10 (Hits@10); (iii) mean reciprocal rank (MRR); and (iv) mean rank (MR). Furthermore, we adopted the filtered rankings \cite{TransE}, which means that we only keep the current testing entity during ranking. Due to DSKG is capable of predicting relations only given one entity, we reported the so-called ``\textbf{cascade}" results. Given a testing triple $(s, r, o)$, DSKG first predicts the relations of $(s, ?)$ to obtain the rank of $r$, and then predicts the entities of $(s, r, ?)$ to obtain the rank of $o$. Finally, these two ranks are multiplied for comparison (i.e., the worst rank).

The experimental results on FB15K and WN18 are illustrated in Table \ref{tab:ent}. Because these two datasets contain many inverse triples, InverseModel, which only learns inverse relations, still achieved competitive performance. Additionally, we can  observe that DSKG outperformed the other models for many metrics. Particularly, DSKG achieved the best performance for Hits@1, which showed that DSKG is quite good at precisely learning to predict entities. Even we evaluated DSKG in the cascade way, it still achieved comparable results.

Table \ref{tab:ent-237} shows the entity prediction results on FB15K-237. We observed that: (1) The performance of all the models slumped. Specifically, InverseModel completely failed on this dataset, which reveals that all the models cannot directly improve their performance by using inverse relations any more. (2) DSKG significantly outperformed the other models for all the metrics. DSKG (cascade) also achieved state-of-the-art performance for some metrics (e.g., Hits@10).

\subsection{Triple Prediction}
\label{subsect:triple prediction}

DSKG is capable of not only predicting entities, but also predicting the whole triples. To evaluating the performance of DSKG on predicting triples directly, we constructed a beam searcher with a large window size. There also exist some complex methods that can improve sequence prediction performance\cite{DeepReinforcementLearning}. Specifically, the model was first asked to take all the entities as input to predict relations, and then the top-100K $(entity, relation)$ pairs were selected to construct the incomplete triples like $(s, r, ?)$. Next, the model took these incomplete triples as input to predict their tail entities. Finally, we chose the top-1M triples as output, and sorted them in descending order for evaluation.

\begin{table*}[pt]  
	\begin{floatrow}  
		\capbtabbox{
			\scriptsize \begin{tabular}{lcccc}
				\toprule
				\ Models\ &\ Hits@1\ &\ Hits@10\ &\ MRR\ &\ MR\ \\ \midrule
				\ TransE$^\dagger$\ & 13.3 & 40.9 & 22.3 & 315 \\
				\ TransR$^\dagger$\ & 10.9 & 38.2 & 19.9 & 417 \\
				\ PTransE$^\dagger$\ & 21.0 & 50.1 & 31.4 & 299 \\ \midrule
				\ DISTMULT\ & 15.5 & 41.9 & 24.1 & 254 \\
				\ NLFeat\ & - & 41.4 & 27.2 & - \\
				\ ComplEx\ & 15.2 & 41.9 & 24.0 & 248 \\
				\ NeuralLP\ & - & 36.2 & 24.0 & - \\
				\ ConvE\ & 23.9 & 49.1 & 31.6 & 246 \\
				\ InverseModel\ & 0.4 & 1.2 & 0.7 & 7,124 \\ \midrule
				DSKG (cascade) & 20.5 & 50.1 & 30.3 & 842 \\
				DSKG & \textbf{24.9} & \textbf{52.1} & \textbf{33.9} & \textbf{175} \\
				\bottomrule
		\end{tabular}}{  
			\caption{Entity prediction on FB15K-237}
			\label{tab:ent-237}
		}  
		\capbtabbox{  
			\scriptsize \begin{tabular}{lccc}
				\toprule
				\ Datasets\ &\ FB15K\ &\ WN18\ &\ FB15K-237\ \\ 
				\midrule
				\ \#Entities     &  14,951 & 40,943 & 14,541 \\
				\ \#Relations  &   1,345 &    18 &  237 \\
				\ \#Train 	& 483,142 & 141,442 &  272,115 \\
				\ \#Valid.   &  50,000 & 5,000 &  17,535 \\
				\ \#Test		&  59,071 & 5,000 &  20,466 \\
				\bottomrule
		\end{tabular}}{  
			\caption{Dataset statistics}
			\label{tab:dataset}
		}
	\end{floatrow}  
\end{table*}

We used precision to assess these output triples. Let $\mathbf{S}_{out}^n$ denote the set of top-$n$ output triples, $\mathbf{S}_{corr}$ denote the set of all correct triples (including the testing, validation and training sets) for a KG, and $\mathbf{S}_{pred}$ denote the set of predicted triples (including the testing and validation sets). The precision $p_n$ w.r.t. top-$n$ output triples is calculated as follows:
\begin{alignat}{2}
n_{corr} &= |\mathbf{S}_{out}^n \cap \mathbf{S}_{corr}|,& \quad n_{pred} &= |\mathbf{S}_{out}^n \cap \mathbf{S}_{pred}|, \notag \\
n_{error} &= |\mathbf{S}_{out}^n| - n_{corr}, &	\quad  p_n &= \frac{n_{pred}}{n_{pred} + n_{error}}, 
\label{eq:tp}%
\end{alignat}
where $n_{corr}, n_{pred}, n_{error}$ denote the correct, predicted, and error numbers in $\mathbf{S}_{out}^n$, respectively. As a result, we can draw the curve of $p_n$ in terms of $n$. 

We conducted experiments on the three datasets, and compared DSKG with two general models: \textbf{G2} and \textbf{G4}. G2 is a general 2-layer RNN model (see Fig.~\ref{fig:methods}c). G4 is a general 4-layer RNN model, as DSKG uses four different RNN cells. All the features (sampler, dropout, etc.) in DSKG were also applied to them.

\begin{figure}[pt]
	\centering
	\includegraphics[width=.98\columnwidth]{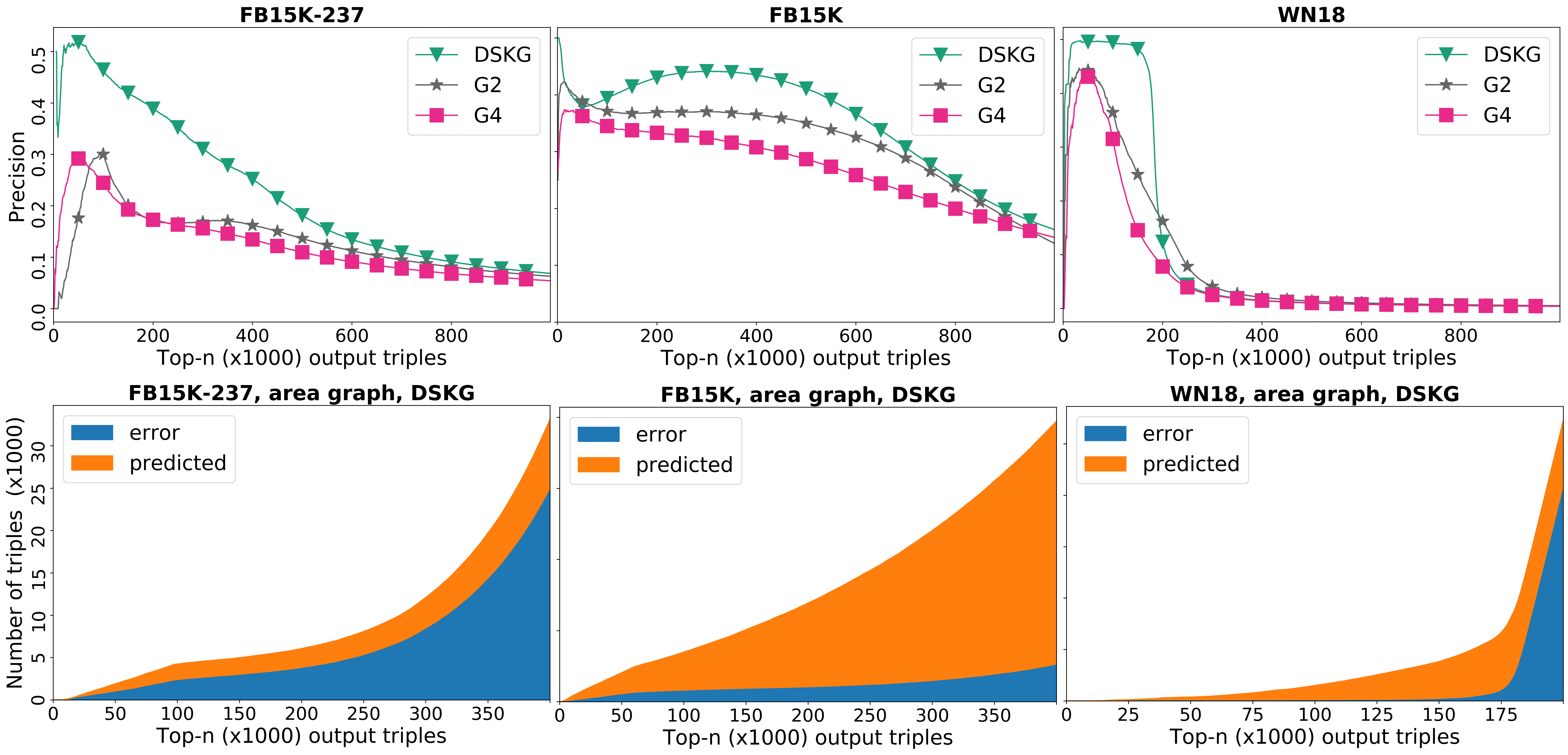}
	\caption{Triple prediction results on three datasets}
	\label{fig:tp}
\end{figure}

As shown in the left column of Fig.~\ref{fig:tp}, DSKG significantly outperformed G2 and G4 on all the datasets, especially for FB15K-237. Also, G4 performed worse than G2. This may be due to that deeper networks and more parameters make the entity and relation embeddings improperly trained. The right column of Fig.~\ref{fig:tp} shows the detailed triple prediction proportions of DSKG. DSKG predicted more than 2,000 correct triples with precision 0.47 (top-100K) on FB15K-237. On the other two easier datasets, DSKG performed better. It correctly predicted 34,155 triples on FB15K with precision 0.87 (top-400K) and 5,037 on WN18 with precision 0.91 (top-170K). Note that the precision of DSKG rapidly sharply dropped on WN18 in the end, because WN18 only has 10,000 triples to predict, while DSKG already output all the triples that it can predict.

\section{Analysis}
\label{sect:analysis}

\subsection{Comparison with Alternative Models}
\label{subsect:alternative}

To analyze the contribution of each part in DSKG, we developed a series of sub-models only containing partial features:

\begin{figure}[t]
	\centering
	\includegraphics[width=.75\columnwidth]{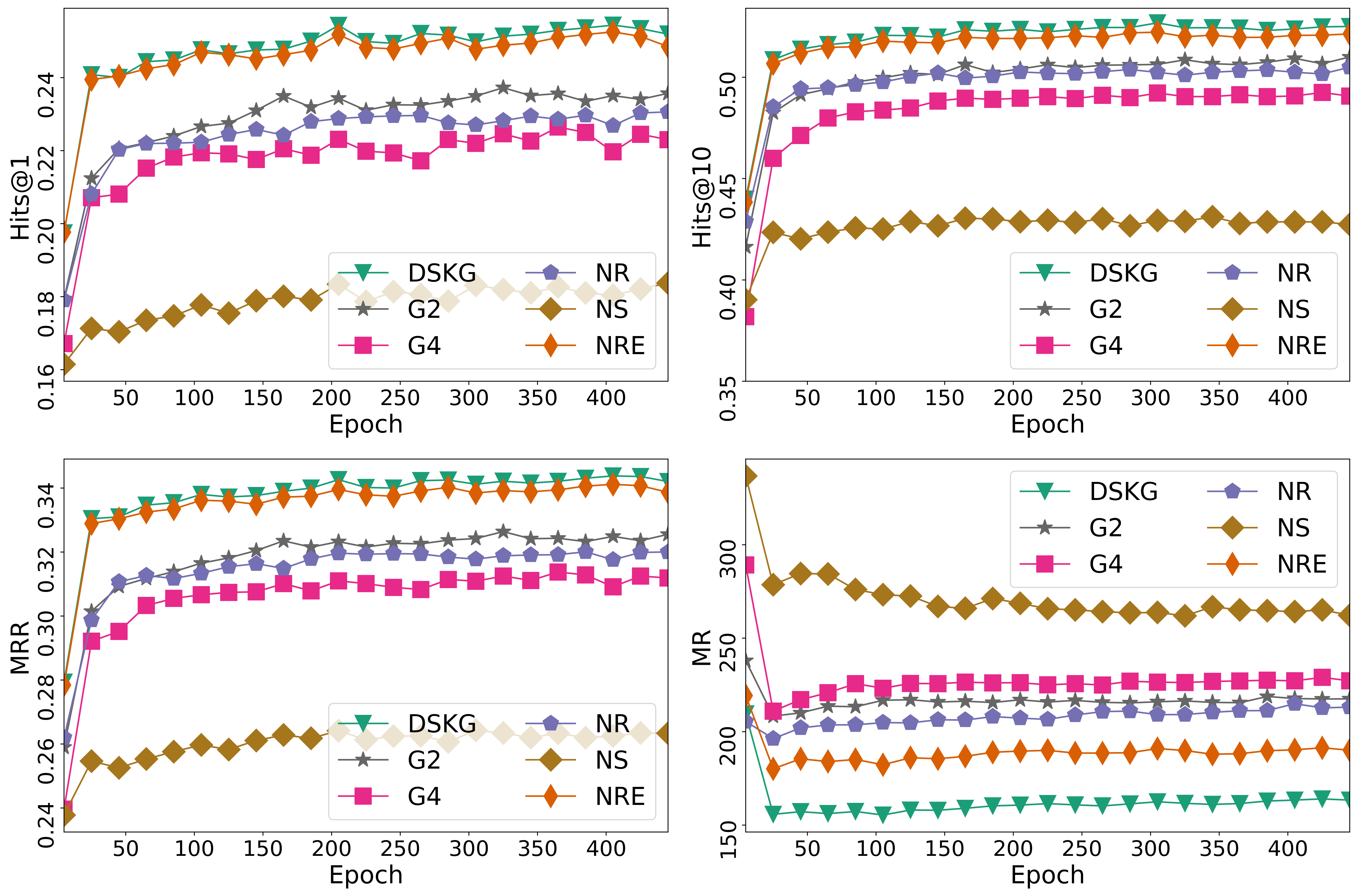}
	\caption{Results of alternative models on FB15K-237}
	\label{fig:pstudy}
\end{figure}

\begin{itemize}
	\item \textbf{NR}. DSKG without using the relation loss in training. We constructed this model to assess the effect of minimizing relation loss to entity prediction.
	\item \textbf{NS} (non-sequential). We used four fully-connected layers (ReLU as the activation function) to replace RNN cells in DSKG, and cut down the connections between relation layer and entity layer. In the end, we added a dense layer to combine the output of these two layers. This model also shares the other features of DSKG (dropout, sampler, etc.). We constructed it to investigate the effect of sequential characteristic.	
	\item \textbf{NRE}. DSKG without using the enhancement method (Section~\ref{subsect:enhance}).
\end{itemize}

Fig.~\ref{fig:pstudy} shows the performance of these models as well as G2 and G4 on the validation set of FB15K-237. From the results, we observed that: 

\begin{itemize}
	\item Sequential characteristic is a key point to DSKG. Comparing DSKG with NR and NS, we can find that: (1) although NR kept the sequential structure, it still performed worse than DSKG, since NR did not learn to predict relations; (2) NS did not use the sequential structure and not learn to predict relations. Hence, it obtained the worst result in Fig.~\ref{fig:pstudy}.
	\item The KG-specific multi-layer RNN architecture significantly improved the performance. DSKG outperformed G2 and G4 for all the metrics on FB15K-237, even it did not use the enhancement from relation prediction. Note that, in Section \ref{subsect:triple prediction}, we have already shown that DSKG performed better than G2 and G4 on triple prediction. Therefore, the architecture used in DSKG has a better capability to model KGs than the general multi-layer RNN models.  
	\item The relation enhancement method further refined the entity prediction results. DSKG always performed better than NRE, especially for MR, since it can directly eliminate many incorrect entities.
\end{itemize}

\subsection{Influence of Layer Number}

We conducted an experiment to analyze the influence of layer number to entity prediction. As shown in Table \ref{tab:layers}, when we increased the layer number from 1 to 4, DSKG cannot continuously improve the performance. The 4-layer model also took more time than the 2-layer model for each epoch (18.7s vs. 13.4s). Thus, we chose the 2-layer DSKG as the main reported version, which achieved the best and was trained quite fast (about one hour using a GTX 1080Ti).

\begin{table*}[t]  
	\begin{floatrow}  
		\capbtabbox{  
			\scriptsize\begin{tabular}{lcccc}
				\toprule
				\ \multirow{2}{*}{No.}\ & \multicolumn{2}{c}{DSKG} & \multicolumn{2}{c}{DSKG (cascade)} \\
				\cmidrule(lr){2-3} \cmidrule(lr){4-5}  &\ Hits@1\ &\ Hits@10\ &\ Hits@1\ &\ Hits@10\ \\ \midrule
				\ 1 & 24.6 & 51.2 & 20.3 & 48.8 \\
				\ 2$^\ddagger$ & \textbf{24.9} & \textbf{52.1} & \textbf{20.5} & \textbf{50.1} \\
				\ 3 & 24.6 & 51.4 & 19.8 & 49.3 \\
				\ 4 & 24.1 & 50.3 & 18.6 & 48.1 \\
				\bottomrule
				\multicolumn{5}{l}{``$\ddagger$" denotes the main results reported in} \\
				\multicolumn{5}{l}{ Section~\ref{sect:exp}.} \\
			\end{tabular}  
		}{  
			\scriptsize\caption{Entity prediction results with varied layer numbers on FB15K-237}
			\label{tab:layers}
		}  
		\capbtabbox{  
			\scriptsize\begin{tabular}{lcccc}
				\toprule
				\ \multirow{2}{*}{Size}\ & \multicolumn{2}{c}{DSKG} & \multicolumn{2}{c}{DSKG (cascade)} \\
				\cmidrule(lr){2-3} \cmidrule(lr){4-5} &\ Hits@1\ &\ Hits@10\ &\ Hits@1\ &\ Hits@10\ \\ \midrule
				\ 512$^\ddagger$ & \textbf{24.9} & \textbf{52.1} & \textbf{20.5} & \textbf{50.1} \\
				\ 256 & 24.8 & \textbf{52.1} & 19.9 & 49.8 \\
				\ 128 & 24.5 & 51.5 & 19.0 & 48.5 \\
				\ 64 & 23.1 & 48.6 & 17.1 & 45.1 \\
				\bottomrule
			\end{tabular} 
		}{  
			\scriptsize\caption{Entity prediction results with varied embedding sizes on FB15K-237}
			\label{tab:embedding-size} 
		}
	\end{floatrow}  
\end{table*}  

\subsection{Influence of Embedding Size}

DSKG is a parameter-efficient model. Table \ref{tab:embedding-size} shows the entity prediction results of DSKG with varied embedding sizes. When we decreased the embedding size to 128, DSKG can still achieve state-of-the-art performance on FB15K-237. When the embedding size was set to 64, which is a very small value, it was also competitive. For DSKG (cascade), due to involving the relation prediction results, the performance decreased more severely, but still acceptable. We chose the embedding size 512 as the main reported version to obtain the best performance.

\section{Conclusion and Future Work}
\label{sect:concl}

In this paper, we proposed a new model to use a KG-specific multi-layer RNN to model triples in a KG as sequences. Our experimental results on three different datasets showed that our models achieved promising performance not only on the traditional entity prediction task, but also on the new triple prediction task. For the future work, we plan to explore the following two directions:

\begin{itemize}
	\item Integrating the attention mechanism \cite{Attention} in our model. While the attention mechanism has shown its power in the NLP area, applying this mechanism to KG completion has not been well studied. In the future, we want to extend DSKG with this mechanism to improve its inference ability.
	\item Using a provided KG to complete another KG. Recently, several methods start to leverage extra textual data for improving KG completion. However, textual data are written in natural language. Due to the ambiguity and heterogeneity, they may bring mistakes in prediction. Therefore, we think that taking existing KGs as another kind of extra data may improve performance. 	
\end{itemize}

\noindent\textbf{Acknowledgements.} This work was supported by the National Natural Science Foundation of China (Nos. 61872172 and 61772264).

%
% ---- Bibliography ----
%
\bibliographystyle{splncs04}
\bibliography{references}

\end{document}